\documentclass[sigconf]{acmart}
\usepackage{graphicx}
\usepackage{algorithm}
\usepackage{algorithmic}
\DeclareMathOperator*{\argmax}{arg\,max} 

\AtBeginDocument{%
  \providecommand\BibTeX{{%
    \normalfont B\kern-0.5em{\scshape i\kern-0.25em b}\kern-0.8em\TeX}}}

\acmConference[Toronto '20]{Toronto '20: ACM Conference on Health, Inference, and Learning}{June 02--04, 2020}{Toronto, Canada}
\acmBooktitle{Toronto '20: ACM Conference on Health, Inference, and Learning, April 02--04, 2020, Toronto, Canada}

\begin{document}

\title[Improving Medical Label Quality With Stratified Noisy Cross-Validation]{Improving Medical Annotation Quality to Decrease Labeling Burden Using Stratified Noisy Cross-Validation}

\author{Joy Hsu}
\authornote{Both authors contributed equally to this research.}
\authornote{Work done while employed at Google.}
\affiliation{\institution{Stanford University}}
\email{joycj@stanford.edu}

\author{Sonia Phene}
\authornotemark[1]
\affiliation{\institution{Google}}
\email{sphene@google.com}

\author{Akinori Mitani}
\affiliation{\institution{Google}}
\email{amitani@google.com}

\author{Jieying Luo}
\affiliation{\institution{Google}}
\email{jieying@google.com}

\author{Naama Hammel}
\affiliation{\institution{Google}}
\email{nhammel@google.com}

\author{Jonathan Krause}
\affiliation{\institution{Google}}
\email{jkrause@google.com}

\author{Rory Sayres}
\affiliation{\institution{Google}}
\email{sayres@google.com}

\begin{abstract}
As machine learning has become increasingly applied to medical imaging data, noise in training labels has emerged as an important challenge. Variability in diagnosis of medical images is well established; in addition, variability in training and attention to task among medical labelers may exacerbate this issue.

Methods for identifying and mitigating the impact of low quality labels have been studied, but are not well characterized in medical imaging tasks. For instance, Noisy Cross-Validation splits the training data into halves, and has been shown to identify low-quality labels in computer vision tasks through comparing model predictions and labels; but it has not been applied to medical imaging tasks specifically. In addition, there may be concerns around label imbalance for medical image sets, where relevant pathology may be rare.

In this work we introduce \textit{Stratified Noisy Cross-Validation} (SNCV), an extension of noisy cross validation. SNCV can provide estimates of confidence in model predictions by assigning a quality score to each example; stratify labels to handle class imbalance; and identify likely low-quality labels to analyse the causes. Sample selection for SNCV occurs after training two models, not during training, which simplifies application of the method.

We assess performance of SNCV on diagnosis of glaucoma suspect risk (GSR) from retinal fundus photographs, a clinically important yet nuanced labeling task. Using training data from a previously-published deep learning model, we compute a continuous quality score (QS) for each training example. We relabel 1,277 low-QS examples using a trained glaucoma specialist; the new labels agree with the SNCV prediction over the initial label >85\% of the time, indicating that low-QS examples mostly reflect labeler errors. We then quantify the impact of training with only high-QS labels, showing that strong model performance may be obtained with many fewer examples. By applying the method to randomly sub-sampled training dataset, we show that our method can reduce labelling burden by approximately 50\% while achieving model performance non-inferior to using the full dataset on multiple held-out test sets.

\end{abstract}


\maketitle

\section{Introduction}
 Machine learning has increasingly been applied to medical imaging tasks, where there are challenges in obtaining high quality data and a strong need for performance robustness.  Medical data can be difficult to annotate and diagnoses can vary amongst clinicians \cite{early1991grading, scott2008agreement, li2010digital, gangaputra2013comparison, ruamviboonsuk2006interobserver, elmore1994variability, elmore2015diagnostic}. Training a robust, generalizable machine learning model requires consideration of this label variability, particularly since most deep learning models have the ability to memorize all the training examples \cite{zhang2016, arpit2017}. Techniques to improve label quality, such as adjudication by multiple graders, do exist but are costly and are therefore used primarily for tune or test sets rather than training sets \cite{krause2018}. Labels graded by single graders, especially less experienced graders, can be less consistent.

 These challenges can be magnified for ambiguous medical issues, such as detecting glaucoma suspect risk (GSR). Glaucoma is a disease that has no single set of established guidelines and relies on more of a `gut feel' assessment. Glaucoma diagnosis consists of the compilation of clinical exam data with functional and structural testing by examiners, but how to combine such evidence for the final diagnosis often relies on clinicians. This ambiguity is exacerbated when only retinal fundus images are available, such as in teleretinal screening settings, where patients may be referred to specialists based only on potential diagnostic features present in the images. Diagnostic criteria for teleretinal programs vary between programs and the link between features and diagnosis is not well established. This introduces a challenge in collecting labels to develop an algorithm to detect GSR from fundus images \cite{thomas2014effectiveness, phene2019deep}.

 Labels collected for ambiguous tasks inherently have errors. Various methods have been proposed to improve performance when training models under the presence of label noise, but they have not been discussed or explored in medical imaging tasks in depth. One challenge in applying such methods to a medical imaging task is the existence of class imbalance. For example, most patients coming to a glaucoma screening program do not have glaucoma, and only a few patients would have glaucoma.
 
 In this study, we adapt Noisy Cross-Validation \cite{chen2019understanding} to select a fixed number of samples from each label (negative and positive) to account for class imbalance, in addition to assigning a novel \textit{Quality Score} (QS) as signal for label noise based on model confidence. We term this approach \textit{Stratified Noisy Cross-Validation} (SNCV), and apply it to a training set from a previously-described deep learning model of GSR \cite{phene2019deep}.
 
 Our primary questions for this study are: \textbf{(1)} Do low-quality labels identified by SNCV reflect actual errors in clinical diagnosis? \textbf{(2)} Using SNCV, can we quantify the impact of low-quality labels on performance of models? And \textbf{(3)} Can SNCV potentially reduce the labeling requirements for modeling efforts?
 
 We address \textbf{(1)} by re-labeling examples with a low QS using a trained glaucoma specialist, and assessing how often the relabeled examples diverge from the initial labels. We address \textbf{(2)} by training models on different subsets of the original training set, using increasing numbers of only high-QS or low-QS examples; and measuring the resulting model performance against an adjudicated, high-quality tune set. We address \textbf{(3)} by repeating the SNCV process on a randomly-selected subset of our original training set, approximately half the original size. We train a model with all examples in the smaller subset, as well as a model with only high-QS examples from the smaller subset; and compare the performance of both models, as well as a model trained on the entire data set, on three independent, held-out test sets. If SNCV is effective at mitigating low-quality labels, we would expect to see a model trained on only high-QS labels outperform one trained on all labels, and potentially be comparable to a model trained with a larger overall data set.

\section{Related Work}

 Among methods to train models under the presence of label noise, several methods require estimating the noise transition matrix during training \cite{sukhbaatar2014training, patrini2017making, goldberger2016training}; but the estimation may be computationally complex. Furthermore, the noise transition matrix approach assumes that the noise probability is based only on the example's true class, although labelers may have different operating points, and the noise may not be able to simply explained by a single noise transition matrix.

 Another approach is to use model predictions during training for label modification or correction \cite{reed2014training, tanaka2018joint, ma2018dimensionality}, sample selection \cite{jiang2017mentornet, han2018co, chen2019understanding, yu2019does},  sample weighting \cite{ren2018learning}, or some combination of these methods \cite{song2019selfie}. Most of these methods are based on the observation that deep learning models can learn simple patterns first before memorizing all the training examples \cite{zhang2016, arpit2017}.

 Here, we focus on Co-teaching\cite{han2018co} and Noisy Cross-Validation\cite{chen2019understanding}, which both train multiple models as part of determining label quality.

\subsection{Co-teaching}

 Co-teaching \cite{han2018co} trains two models, during which each model ``teaches'' the other. For each mini-batch used during training, each model selects $R(t)$  small loss instances in for the other model to update its weights.  $R(t)$ represents a dynamic schedule for how many samples to select, and the authors propose to start from $R(t)=1$ and gradually decrease it to $1 - \epsilon$ over 10 epochs, given the known noise level $\epsilon$. Yu et al \cite{yu2019diagreement} argued that the two network may converge eventually and suffer from accumulated error issue caused by sampling bias. This suggests that if the two networks tend to have bias in the same direction, this method may further reinforce the bias. This can cause issues in case of class imbalance, as the models tend to have small loss for examples in the majority class especially at the beginning of the training.
 
 One possible extension is to select $R(t)$ samples separately from each class. However, co-teaching selects samples based on the training loss in a mini-batch, and each mini-batch may contain only one or two cases from the minor class. This makes it difficult to keep certain proportion of the minor class examples.

\subsection{Noisy Cross-Validation}

 Chen et al introduced Noisy Cross-Validation (NCV) as an initial step to select mostly correct examples before co-teaching \cite{chen2019understanding}. This method splits the dataset randomly into two subsets, trains two models on the subsets, and selects samples based on matching predictions by the model trained with the other subset. NCV was shown to be able to identify examples that are likely to have wrong labels \cite{chen2019understanding}. In addition, this method separates sample selection from model training, making adapting and extending this method simpler. 
 
\section{Methods}

\subsection{Data sources}
 All images were de-identified according to HIPAA Safe Harbor prior to transfer to the study investigators. Ethics review and Institutional Review Board exemption were obtained using Quorum Review Institutional Review Board. All research complied with the Health Insurance Portability and Accountability Act of 1996.

 The development (train and tune) datasets consisted of fundus images obtained from multiple sources: EyePACS \cite{cuadros2009eyepacs}, Inoveon \cite{inoveonwebsite} and the Age-Related Eye Disease Study \cite{age1999age} in the United States; and three eye hospitals in India (Aravind Eye Hospital, Sankara Nethralaya, and Narayana Nethralaya). The development dataset was further split into the train dataset (70,000 fundus images) and the tune dataset (1,508 images). The train dataset was enriched for GSR examples using active learning \cite{settles2012active} and was single graded by 43 graders who had reviewed guidelines and questions for the task. The tune set consisted of 1,508 images, each independently graded by three glaucoma specialists (from a group of 11 specialists).
	
 Three datasets were used for validation. Test set ``A'' comprised a random subset of color fundus images from EyePACS, Inoveon, UK Biobank, AREDS, and Sankara Nethralaya. Test set ``B'' comprised macula-centered color fundus images from the Atlanta Veterans Affairs (VA) Eye Clinic diabetic teleretinal screening program. Test set ``C'' comprised macula-centered and disc-centered color fundus images from the glaucoma clinic at Dr. Shroff's Charity Eye Hospital, New Delhi, India. 
	
 From each source, only one image per patient was used in the final analysis, and each image appeared only in one of the train, tune, or test sets.
	
\subsection{Labels}

 For model development and performance analysis, we used labels obtained to train a previously-described deep learning model for glaucoma suspect risk \cite{phene2019deep}. For our research, we focus primarily on \textit{glaucoma suspect risk} (GSR), which is a four-way classification of each image as one of [non-glaucomatous, low risk, high risk, likely glaucoma]. We also consider a binary refer / no-refer decision: cases labeled as non-glaucomatous or low risk would not be referred to a specialist, while high risk and likely glaucoma cases would be referred.
	
 In addition to overall GSR, for each image in the development datasets and test set ``A'', graders were asked to indicate the presence of several lesions often associated with glaucoma. These lesion types were identified from the American Academy of Ophthalmology's Preferred Practice Patterns. These labels are used here to characterize different types of examples, and assess whether specific pathologies are associated with label quality. 
	
 The training dataset was labelled by a total of 43 graders (14 fellowship-trained glaucoma specialists, 26 ophthalmologists, and 3 optometrists) trained on specific guideline and certified for the task. The training set was single graded. 
	
 The tune dataset was triple-graded by fellowship-trained glaucoma specialists (out of a pool of 11 graders). Each grader independently assessed each image.
	
 The labels in the test set ``A'' were determined by adjudication of a rotating panel of 3 fellowship-trained glaucoma specialists (out of a pool of 12 graders). Each grader independently graded an image. Then, in random order, each grader was able to see their own and others' grades for an image and change their answers. Adjudication was run till each grader had a total of two passes at this image. In cases where disagreement persisted, the median value of the three grades was taken, which is equivalent to majority vote in case of a two-way agreement. For validation datasets ``B'' and ``C'', the labels were based on the data provided by the partner organization, as described in the Supplementary Materials.
	
 For the relabeling experiment, a glaucoma specialist assessed each image without access to the previous labels and without the knowledge that examples were selected for relabeling based on having a low quality score. The specialist was independent of the original labelers, and was trained in our internal labeling guidelines. The grading task presented was identical to previous ones, meaning the grader assessed the images for several lesions as well as overall GSR.
	
\subsection{Model training}
\subsubsection{Architecture}

 Our model architecture uses an Inception V3 model pre-trained on the ImageNet dataset, similar to that described by \cite{gulshan2016development}. Color fundus images were resized to 587x587 resolution for input, and the models output softmax scores for each possible value in each multiclass prediction. The model was trained with 17 heads total, including the four-class GSR head, as well as heads for individual glaucoma-associated pathologies. 

\subsubsection{Baseline model}

 As a baseline for comparison, models were trained using all the examples in the train set. To prevent overfitting, the training was terminated before convergence using early stopping based on the performance of referable GSR task on the tune set. Hyper-parameters used are detailed in Supplementary Materials.

\subsection{Stratified Noisy Cross-Validation (SNCV)}

 We extend Noisy Cross-Validation without iteration \cite{chen2019understanding} to assign a quality score to each train example based on model confidence, and support label-based stratification during sample selection in case of class imbalance.

 Chen et al introduced noisy cross-validation as a pre-step before co-teaching to select examples that are likely correct \cite{han2018co}. Their approach splits the dataset randomly into two subsets, trains two models on the subsets, and selects samples from a subset based on matching predictions.

 The original implementation of NCV selects examples with matching model predictions and initial labels -- examples we may refer to as high-confidence examples. One potential concern about this approach is that, if the rate of high-confidence examples varies between positive and negative cases, this approach may exacerbate class imbalance. For instance, for a medical pathology there may be many patients who can be easily determined not to have the pathology, but a relatively smaller rate of patients with the pathology who can be easily diagnosed. In such a situation, NCV may tend to over-select negative cases, exacerbating an already-existing class imbalance in which positive cases may be rare, as with many medical imaging applications \cite{mazurowski2008training}. With fewer positive examples to train on, NCV without stratification may increase the risk of overfitting, or otherwise degrade model performance.

 Motivated by these concerns, we propose \textit{Stratified Noisy Cross-Validation} (SNCV), a novel approach that assigns each example in the train set a \textit{Quality Score} (QS). In contrast to the original implementation of NCV, our QS accounts for a degree of agreement between the cross-validation model predictions and the initial label, allowing for a graded measure of confidence.

\begin{figure}[htb]
  \includegraphics[width=\linewidth]{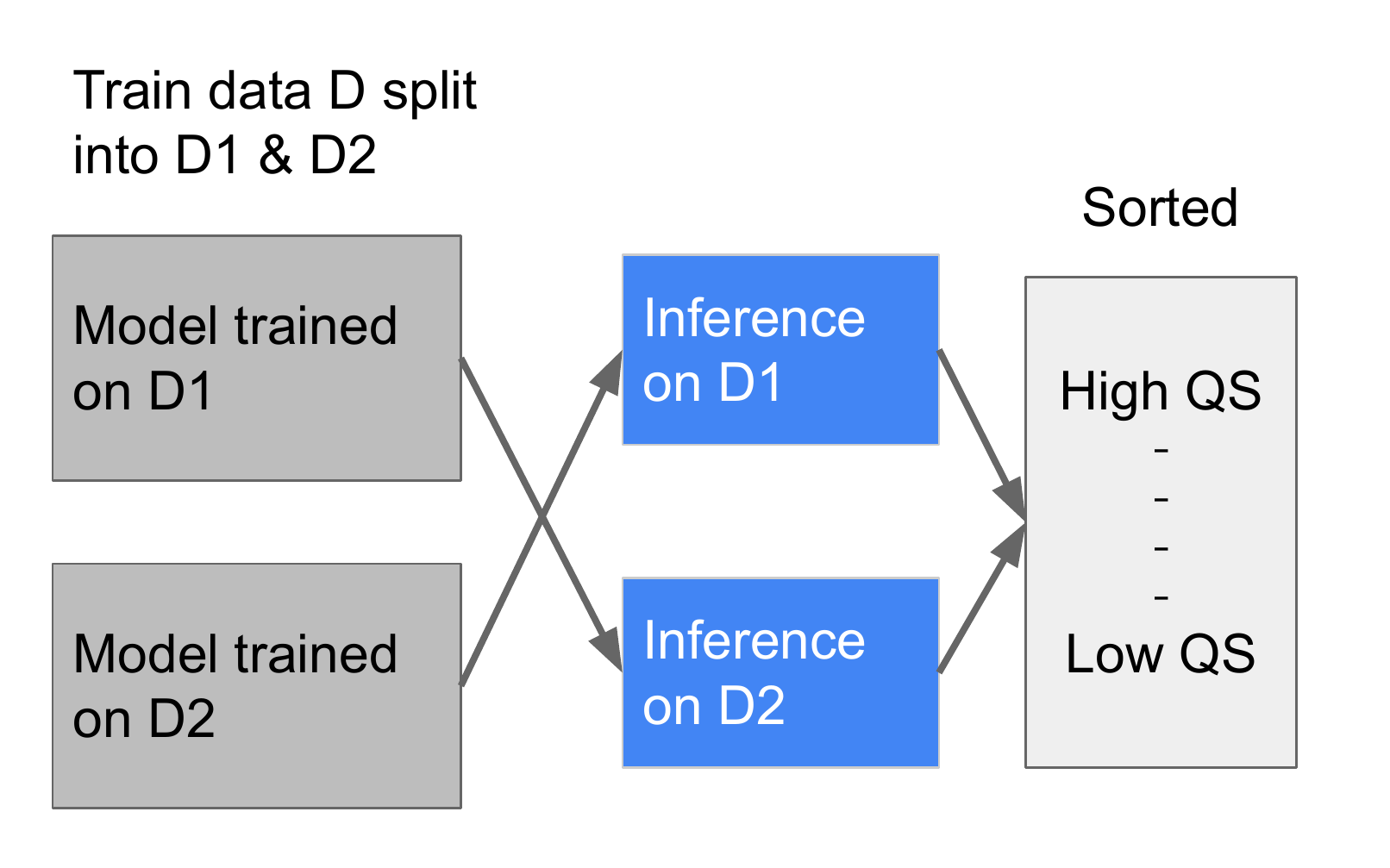}
  \caption{Schematic of stratified noisy cross-validation (SNCV).}
  \label{fig:ncv}
\end{figure}

 In SNCV, the train set $D$ is randomly split equally into two disjoint parts, $D1$ and $D2$. This split allows us to train two models, $M1$ and $M2$, on the respective datasets using the model architecture described in $3.3.1$. We then run inference on the separate, uncontaminated splits -- applying $M1$ to $D2$ and $M2$ to $D1$ -- in order to determine the quality score. Figure \ref{fig:ncv} illustrates this process. 
 
Since $M1$ and $M2$ were based on random splits, we verified these models are comparable to each other by evaluating them on the tune set. They came within 2\% of correct performance. For referable GSR, $M1$ was correct for 45\% of examples and $M2$ was correct 43\%. For non-referable cases, $M1$ and $M2$ achieved 90\% and 91\% correct percentages, respectively. 

 In our work, \textit{Quality Score} (QS) is determined by model softmax scores and alignment between model prediction and label. We define QS as follows: Let $K$ denote a set of all class labels, and $K_+ (\subset K)$ denote a set of class labels corresponding to positive cases.  $x_j$ denotes an output logit from the model for a class $j \in K$. $i$ be the index of the largest output ($= \argmax_{j\in K} x_j$). $y$ be the index of the class from the label. Then:

\begin{equation}
    \text{QS}(x, y)=
    \begin{cases}
      \text{if } i \in K_+ \text{ and } y \in K_+ \text{ , or } i \notin K_+ \text{ and } y \notin K_+ \\
      \frac{\exp(x_i)}{\sum_{j \in K} \exp(x_j)}, \\
      \text{otherwise} \\
      -\frac{\exp(x_i)}{\sum_{j \in K} \exp(x_j)}.  \\
    \end{cases}
  \end{equation} \\
 
 In this estimation, the largest softmax output of the model across classes is taken as a signal of model confidence. Disagreements between the initial labels and predictions beyond binary decision boundary by the intermediate models $M1$ and $M2$ are considered lower quality if the model predictions have a higher softmax score.
 
 For instance, consider two possible cases of model/label disagreement: for example A, out of the four possible severities -- non-glaucomatous, low risk, high risk, and likely glaucoma -- the model predicts that the fundus image is high risk for glaucoma with a score of $0.95$; for example B, it predicts high risk with score $0.60$. Suppose in both of these cases, the initial label is no risk.  We may consider both examples as possible low-quality, due to this mismatch between label and model prediction. But we weigh example A as having a proportionately lower quality, due to the higher confidence expressed by the model output.
 
 Note that this approach relies on each half of the data split $D1$ and $D2$ having a sufficient number of examples to train a model with at least moderate generalization ability.

 For the specific case of the four-class GSR head, ``non-glaucomatous'' and ``low risk'' are not included in $K_+$, and ``high risk'' and ``likely glaucoma'' are included in $K_+$. Thus, we only consider label mismatches if they cross the referability boundary: if the initial label suggests referring a case and the model predicts no referral, or \textit{vice versa}. We add this in order to avoid potentially spurious identification of low-quality labels based on disagreement around the degree of pathology observed, rather than the more clinically-relevant criterion of referable pathology.

 When selecting examples with high quality scores, class imbalance has to be taken into account. Glaucoma referable cases are rarer and tend to have lower quality scores; therefore to ensure that the model encounters a sufficient number of positive examples during training, we stratify examples by their initial labels. Examples with positive (high risk and likely glaucoma) and negative (non-glaucomatous and low risk) labels in the train set were ranked by quality scores separately. For each band of quality scores (e.g. highest 1k, 10k, 25k ...), we selected the same proportion of positive and negative cases as we observe in the overall training set. For example, when selecting the 1k examples in total, 236 examples were chosen from the positive cases, and 764 cases were chosen from the negative cases, based on the ranking in each class. The algorithm is summarized in Algorithm \ref{Alg1}.

\begin{algorithm}[t]
	\caption{Stratified Noisy Cross-Validation}
	\label{Alg1}
	\textbf{INPUT:} the noisy set $\mathcal{D}$, the number of examples to choose $k$
	\begin{algorithmic}[1]
	    \STATE Obtain positive label rate $\tau = \frac{\Sigma_{(x, y) \in \mathcal{D}} y}{|\mathcal{D}|}$
	    \STATE Randomly split $\mathcal{D}$ into two halves, $\mathcal{D}_1$ and  $\mathcal{D}_2$
	    \STATE Train a model $M_1$ on $\mathcal{D}_1$
        \STATE Obtain quality scores for $\mathcal{D}_2$ by applying $M_1$ 
        \STATE Train a model $M_2$ on $\mathcal{D}_2$
        \STATE Obtain quality scores for $\mathcal{D}_1$ by applying $M_2$
        \STATE Rank positive and negative examples in $\mathcal{D}$ by quality scores
        \STATE Select $\tau k$ positive high-quality examples from $\mathcal{D}$ ($\mathcal{D}'_{+}$)
        \STATE Select $(1 - \tau) k$ negative high-quality examples $\mathcal{D}$ ($\mathcal{D}'_{-}$)
        \STATE Train a model $M_3$ on $\mathcal{D}'_{+} \cup \mathcal{D}'_{-}$    
	\end{algorithmic}
\textbf{OUTPUT:} the final trained model, $M_3$
\end{algorithm}

\section{Results}

\subsection{Distribution of quality scores}

 The distribution of label quality scores across examples in our training set is plotted in Figure \ref{fig:qs_dist}. We separately plot these distributions based on whether the initial label indicated non-referable GSR (for responses of non-glaucomatous and low risk, top plot) or referable GSR (for high risk or likely glaucoma responses, bottom plot).
 
 Both of the distributions illustrate some common features of the distributions. Quality scores range in [-1, -0.25] and [0.25, 1], due to the way we define the score (multiplying the largest softmax score of 4 classes originally in [0.25, 1] by -1 if the initial label and model prediction diverge on overall GSR). In both sets of the data, we observe prominent modes centered around -0.4 and 0.4; these indicate low-confidence cases in which the SNCV prediction either disagreed with the initial label (around -0.4) or agreed with it (0.4). These modes appear of comparable size; around the same fraction of low-confidence cases reflect low QS as high QS. This applies both to cases initially marked as non-referable or referable (comparable sizes in both plots).
    
 By contrast, the quality score of non-referable examples includes a strong mode of high scores, near +1: These represent cases where both the initial label and SNCV prediction indicate a non-referable case. This mode is mostly absent among cases initially marked as referable. This suggests that the large majority of high-confidence, high-quality labels are for negative cases: This exacerbates the existing class imbalance for GSR among the population. These results indicate that accounting for class imbalance will be an essential aspect of accounting for label quality, motivating the stratification property of SNCV.

\subsection{Relabelling results of low quality labels}
 Quality scores for every example in the train set were identified using SNCV, utilizing GSR as quality metric. After sorting each train example by score, we sent the 1,277 cases with the lowest quality score to be relabelled by a glaucoma specialist in order to verify the validity of the QS. Table \ref{tab:relabel_cm} shows the confusion matrix of relabeling values. 
 
\begin{table}[htb]
  \caption{GSR confusion matrix}
  \label{tab:relabel_cm}
  \begin{tabular}{p{2.5cm}p{2.5cm}p{2.5cm}}
    \toprule
     & Glaucoma specialist non-refer count & Glaucoma specialist refer count \\
    \midrule
    Original non-refer count & 144 & 372\\
    Original refer count & 667 & 94\\
  \bottomrule
\end{tabular}
\end{table}

 During regrading, the glaucoma specialist was able to select any answer for all of the lesions as well as overall GSR, and was not shown the previous grades nor informed that this was a relabeling task. We then compared how many labels were changed from initial grade to regrade. The glaucoma expert agreed with the model over the initial label on GSR for a large majority (85\%) of reviewed cases (\ref{tab:relabel_cm}). This high rate suggest that most of the low-quality labels are due to initial grader errors. The grader was also able to change answers for any of the lesion types. Table \ref{tab:relabel_freq} shows the relabeling rates for all lesions based on low GSR quality score. 

 Each of these features were graded on a scale that had more than two tiers (see \cite{phene2019deep} for exact questions) and were grouped together into a binary `referable' or `non-referable' value for that particularly lesion. For example, the presence of a particular lesion may have values such as `Yes', `Possible', or `No.' This could be binarized into a referability score by grouping `Yes'/`Possible' vs `No'.

 Table \ref{tab:relabel_freq} below describes the relabeling rate: that is, the fraction of examples for which the glaucoma specialist provided a GSR label that differed from the initial label.
    
\begin{table}[htb]
  \caption{GSR relabel rate}
  \label{tab:relabel_freq}
  \begin{tabular}{ccl}
    \toprule
    Lesion & Relabel rate\\
    \midrule
    GSR & 85.49\%\\
    Notch presence & 19.46\%\\
    Disc hemorrhage & 9.14\%\\
    RNFL defect & 34.07\%\\
  \bottomrule
\end{tabular}
\end{table}

 Table \ref{tab:bin_relabel_freq} describes the relabeling rate, considering only if the new label would lead to a different binary referral value.
 
\begin{table}
  \caption{Binarized referable relabel rate}
  \label{tab:bin_relabel_freq}
  \begin{tabular}{ccl}
    \toprule
    Lesion & Binarized relabel rate\\
    \midrule
    GSR & 81.36\%\\
    Notch presence & 19.26\%\\
    Disc hemorrhage & 8.95\%\\
    RNFL defect & 32.84\%\\
  \bottomrule
\end{tabular}
\end{table}

  Tables \ref{tab:relabel_freq} and \ref{tab:bin_relabel_freq} indicate that many of the estimated low-quality labels may have reflected error on the part of the initial grader, and that examples with low-quality scores for GSR may relate to other lesions being mislabeled in the fundus images as well.
    
\begin{figure}[htb]
      \includegraphics[width=\linewidth]{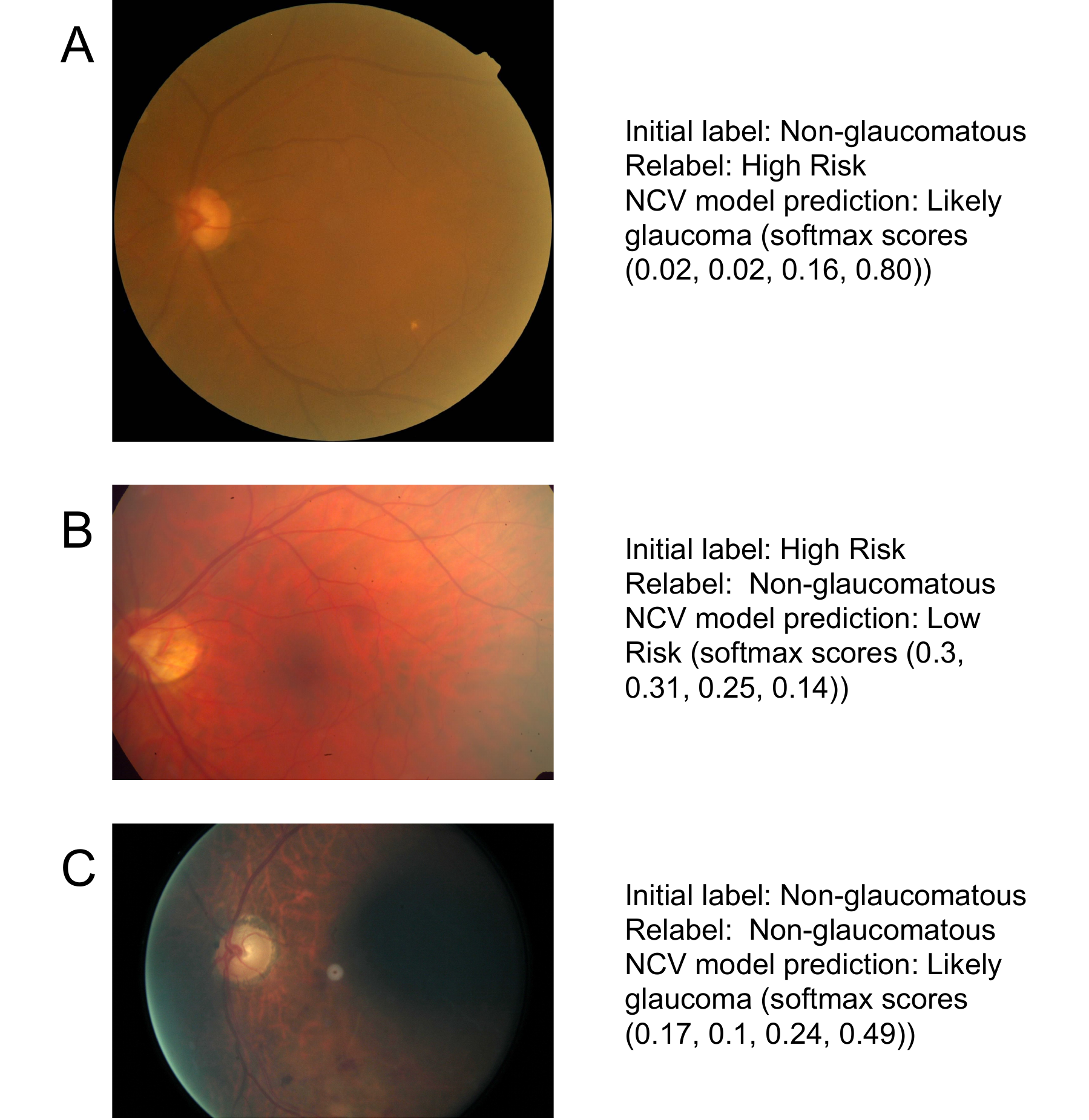}
      \caption{Example fundus images from the relabeling experiment, illustrating different patterns of low quality score. The image in panel b is presented in the original format and has not been zoomed or cropped. The image in panel c has brightness increased slightly to indicate relevant pathology. (During the labeling task, medical labelers were able to manually adjust image brightness and contrast levels.)}
      \label{fig:example_fundus_images}
    \end{figure}

  To gain more insight into what conditions may have resulted in low quality scores, we reviewed selected examples in-person with a second glaucoma specialist. We found some patterns, illustrated in Figure \ref{fig:example_fundus_images}.
    
  The most common pattern we reviewed is reflected in Figure \ref{fig:example_fundus_images}a: in this case there is clear evidence of pathology associated with GSR, which was missed by the initial labeler, but caught by SNCV and the second labeler. In the illustrated case, there is a high cup-to-disc ratio, indicating high GSR risk, but the initial labeler appears not to have recognized it. Whether this reflected a temporary attention lapse, or lack of training, is unclear. 
    
  Figure \ref{fig:example_fundus_images}b and Figure \ref{fig:example_fundus_images}c indicate less-common examples where the cause of low quality score may have been ambiguous or nuanced. Figure \ref{fig:example_fundus_images}b is a case that appears to come from someone with high myopia; this condition causes the optic disc to be tilted, which may have been interpreted by the initial labeler as high GSR risk. This is a case of an uncommon alternate diagnosis which may confound medical labelers without sufficient training or experience. Although we observed many cases in which the initial label was referable GSR, and the relabel was non-referable (Table \ref{tab:relabel_cm}), our qualitative review did not indicate a systematic pattern in the reasons for the seeming false positive in the initial label; rather, the cases we reviewed were idiosyncratic as shown in the illustration. Figure \ref{fig:example_fundus_images}c indicates a case in which the second labeler agreed with the initial grader: both indicated the case is non-glaucomatous, while the SNCV model predicted that this case reflected referable GSR. Secondary review by another glaucoma specialist agreed with the model, based on the appearance of the optic disk and blood vessels. This reflects a highly rare but ambiguous case for which a clear clinical reference standard may not exist, without further patient history.

\subsection{Analysis on other lesions}

 We also considered high and low quality AUC for four lesion types that are of interest for glaucoma: notch, disc hemorrhage, retinal fiber layer (RNFL) defect, and vertical cup-to-disc ratio (VCDR). Of these types, VCDR is the most easy to measure since a ratio of the cup and disc is easiest to observe across image types (some ambiguity still exists for images that are difficult to grade). 

 The same SNCV technique used on referable GSR can be applied to the other lesions. These lesions naturally occur at varying rates in glaucomatous patients. Therefore, we first determined the proportions of referable cases for each of them in our entire training set, which was enhanced using active learning for positive GSR cases. For notch, the positive-negative split is 0.13:0.87, disc hemorrhage is 0.08:0.92, RNFL is 0.13:0.87, and VCDR is 0.17:0.83. Then we applied our same learning technique to identify the QS for images per lesion type. 

 We built models using 25k of the highest QS images and 25k of the lowest QS images, and then ran this model on the tune set. In Table \ref{tab:auc_lesions}, we see the same trend as with GSR, in which the high-quality examples produces high AUC, whereas the low-quality examples produce AUC < 0.5 for notch presence, disc hemorrhage, and RNFL defect. VCDR, the least ambiguous classification and most objective measure on the fundus image, has an expected smallest difference between the high-quality and low-quality examples. Notch presence, disc hemorrhage, and RNFL defect, which are parts of the fundus image that are easily missed and harder to distinguish, shows a significant difference in AUC between models trained from high-QS and low-QS examples. This indicates that quality score can be applied to ambiguous medical classifications to determine correctness of labels.

    \begin{table}[htb]
      \caption{AUC performance based on lesions}
      \label{tab:auc_lesions}
      \begin{tabular}{p{2.2cm}p{2cm}p{2cm}p{1cm}}
        \toprule
        Individual lesion & Highest-quality 25k & Lowest-quality 25k & All data\\
        \midrule
        Notch presence & 0.882 & 0.141 & 0.908\\
        Disc hemorrhage & 0.729 & 0.382 & 0.758\\
        RNFL defect & 0.717 & 0.370 & 0.778\\
        VCDR & 0.926 & 0.882 & 0.922\\
      \bottomrule
    \end{tabular}
    \end{table}

\subsection{Experiments considering grader background}
 We also considered the impact a grader's background may have on the label. Many different graders labeled the training images used - optometrists, ophthalmologists, retina specialists, trainees/fellows, and glaucoma specialists. Glaucoma specialists are most exposed to positive cases of glaucoma and may be more familiar with different ways this anatomy can present in the eye. 

 Given the assigned quality score of train examples from our method, we examined the rates for which a grader disagreed with the models prediction, grouped together by their background. Since we already know that the model may categorize an uncertain positive case as `incorrect', we looked at graders who had more than 30\% of their labels mismatched with the model's predictions. Table \ref{tab:grader_roles} demonstrates that certain backgrounds had more potential `errors.' Glaucoma specialists, for instance, made up 34\% of graders and had 0\% rate of an individual grader with 30\% or more mismatched labels.

\begin{table}[htb]
  \caption{Grader roles and rates of mismatched labels}
  \label{tab:grader_roles}
  \begin{tabular}{p{2.6cm}p{2cm}p{2cm}p{1cm}}
    \toprule
    Role & \% of role in high mismatch rate group &  \% of role in whole group \\
    \midrule
    Glaucoma Specialist & 0\% & 34\%\\
    Retina Specialist & 19\% & 16\%\\
    Ophthalmologist & 25\% & 18\%\\
    Trainee/Fellow & 44\% &  24\%\\
    Optometrist & 12\% & 8\%\\
  \bottomrule
\end{tabular}
\end{table}
    
 Based on this heuristic from SNCV, we attempted to build a model using only glaucoma specialists grades. This had 44k labels from our entire training set and achieved an AUC of 0.941 (CI: 0.924, 0.955).

 Given that glaucoma specialists did grade the bulk of these images, it is also possible that the SNCV method could be biased to their perspective. Therefore, there is a limitation to the claims we can make about mismatched labels. 

 For tasks where specialists may have the most insight, it could be possible to reduce labeling time by simply hiring and only using specialists for the task. However, this does potentially imply increased cost of obtaining the label.

\subsection{SNCV Experiments}
\subsubsection{Importance of stratification}

\begin{figure}[htb]
  \includegraphics[width=\linewidth]{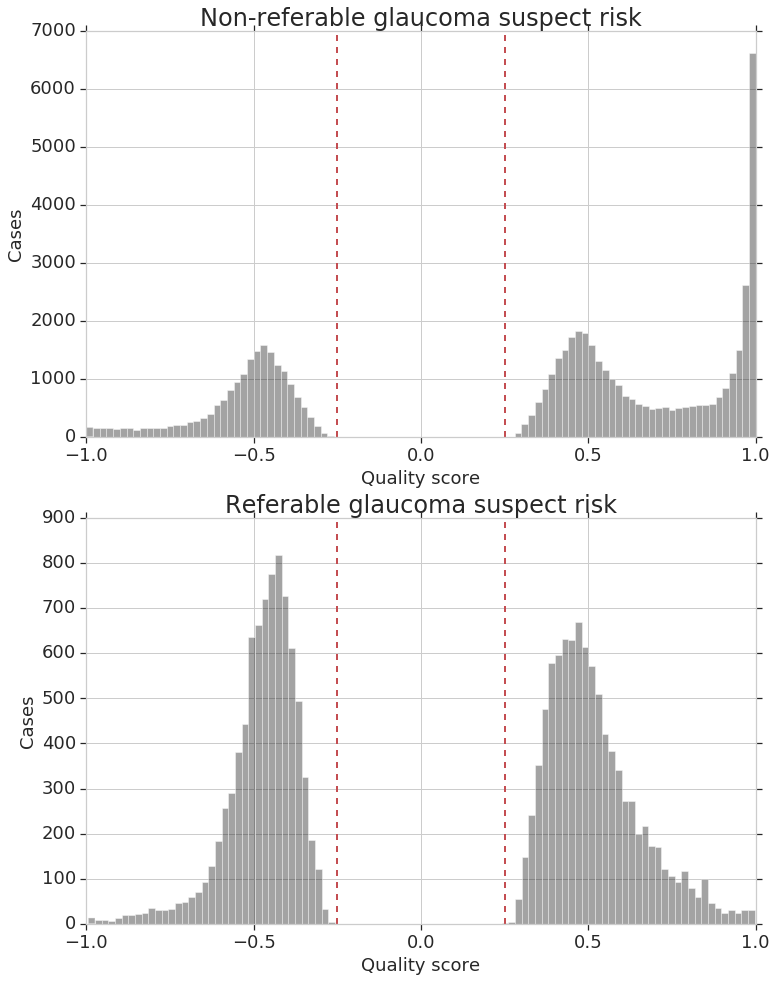}
  \caption{Distribution of quality scores computed with SNCV, for cases initially labeled as non-referable GSR (top) and referable GSR (bottom). Red dotted lines indicate a region of the quality scores that are not populated, due to the method for computing the score; the lowest max value in a 4-class model being 0.25 (see Methods).}
  \label{fig:qs_dist}
\end{figure}

 Using SNCV, we are able to identify high QS and low QS labels. However, for challenging medical diagnoses such as glaucoma, positive referable cases can be ambiguous. There are more obvious negative cases than positive ones (Figure \ref{fig:qs_dist}). Simply taking examples ranked by quality score may overly penalize ambiguous positive cases.

We are also able to see this difference in the distributions themselves of the images with the highest and lowest scores. Table \ref{tab:high_qs_freq} below shows the distribution sorting by highest QS first. The first 1k examples do not contain any positive examples, indicating that the most confident cases are indeed only the negative ones.
\begin{table}[htb]
      \caption{Percentages of referable GSR when sorting by highest quality scores}
      \label{tab:high_qs_freq}
      \begin{tabular}{p{3cm}p{2.2cm}p{2cm}}
        \toprule
        Number of examples & Referable GSR \% &  Non-refer \% \\
        \midrule
        1k & 0\% & 100\%\\
        10k &0.72\% & 99.28\%\\
        25k & 8.52\% & 91.48\%\\
        30k & 10\% & 90\%\\
        35k & 10.72\% & 89.28\%\\
        40k & 10.96\% & 89.04\%\\
        45k & 11.13\% & 88.87\%\\
        70k (All) & 24.6\% & 75.4\%\\
      \bottomrule
    \end{tabular}
    \end{table}

Similarly, Table \ref{tab:low_qs_freq} shows that for the lowest QS labels, the large proportion of them tend to be referable GSR (e.g. more than 60\% in the lowest 10k), which further indicates that the model matching alone may bias selection of high-quality labels. 
\begin{table}[htb]
      \caption{Percentages of referable GSR when sorting by lowest quality scores}
      \label{tab:low_qs_freq}
      \begin{tabular}{p{3cm}p{2.2cm}p{2cm}}
        \toprule
        Number of examples & Referable GSR \% &  Non-refer \% \\
        \midrule
        1k & 37.1\% & 62.9\%\\
        10k & 63.93\% & 36.07\%\\
        25k & 44.05\% & 55.95\%\\
        30k & 38.66\% & 61.34\%\\
        35k & 34.98\% & 65.02\%\\
        40k & 32.51\% & 67.49\%\\
        45k & 30.9\% & 69.1\%\\
        70k (All) & 24.6\% & 75.4\%\\
      \bottomrule
    \end{tabular}
    \end{table}

\subsubsection{Experiments using the full dataset}
\begin{figure}[htb]
  \includegraphics[width=\linewidth]{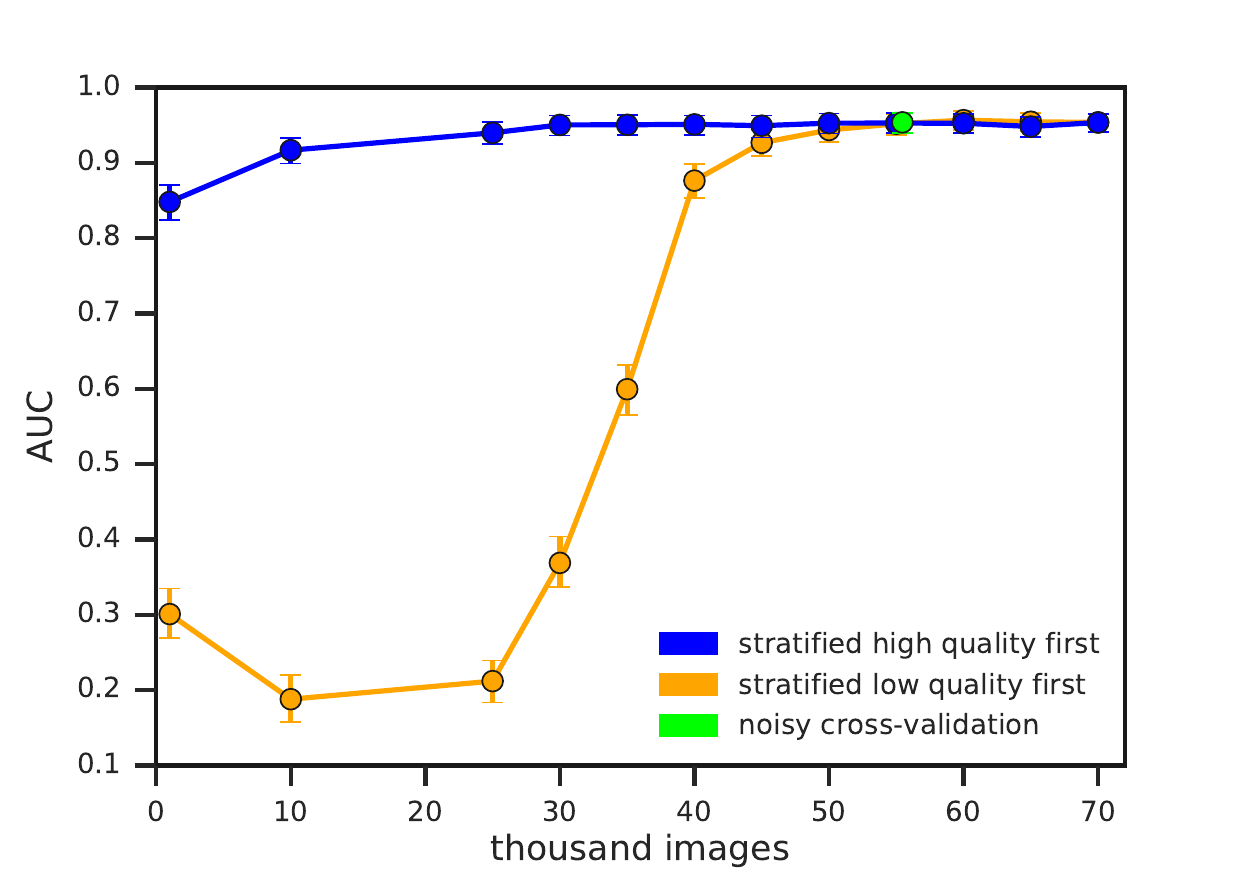}
  \includegraphics[width=\linewidth]{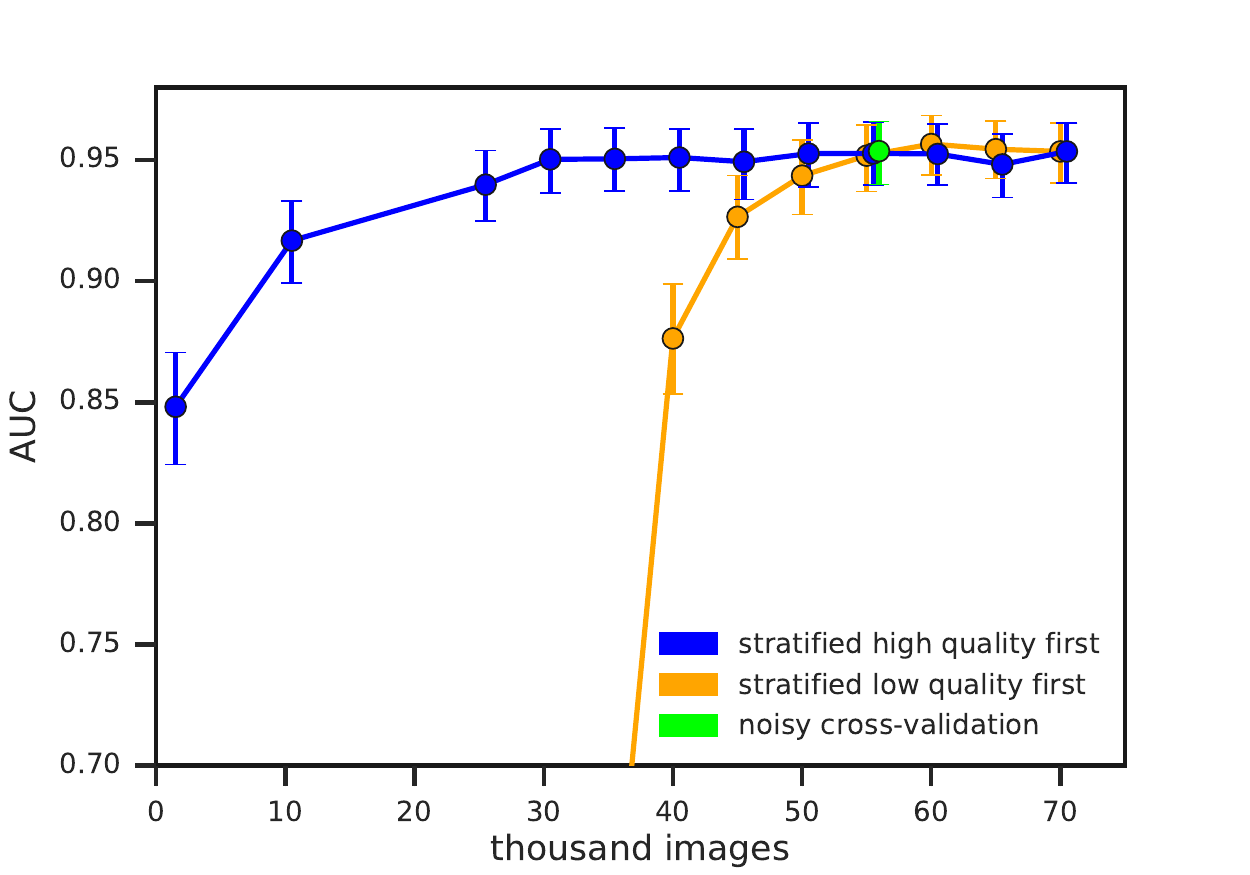}
  \caption{GSR tune set AUC comparison of models trained from high and low quality examples. For each data point, we took the corresponding highest and lowest QS number of samples stratified to include similar proportions of positive GSR cases as the entire dataset. For reference, the data point of using NCV alone is plotted. The two panels are the same except that they are in different scales.}
  \label{fig:ncv_auc_stratified}
\end{figure}

We used the SNCV approach to train models using selected subsets at different sizes from 1k, 10k, 25k, 30k, 35k, 40k, 45k, 50k, 55k, 60k, 65k, as well as the full train set (70k). For each dataset size $n$, we selected the most high quality and the most low quality $n$ examples to train models while including enough positive examples such that the proportion of referable GSR cases was equivalent to the proportion present in the entire 70k set. We then compared the area under the receiver operating characteristic curve (AUC) for each of these intervals on our tune set (Figure \ref{fig:ncv_auc_stratified}). This allowed us to identify and quantitatively measure the effect of selecting examples based on the quality scores on model performance.

The models trained with low quality score labels (up to 30k lowest QS first) reported AUC < 0.5, showing negative correlation between low-quality train data labels and tune labels. Since the tune set was triple graded by glaucoma specialists and the train set was single grading by a host of doctor backgrounds, the performance on the tune set is closer to ground truth and more indicative of the model's performance. The delta between models trained from high-quality and low-quality examples is significant until near the midpoint of splitting train examples.

We also compared this method to using the NCV method using the full train set without stratification. Because NCV only considers the correct examples, this resulted in a single data point, shown in Figure \ref{fig:ncv_auc_stratified}.

Table \ref{tab:auc_highqs_stratified} shows the difference in performance if we took the highest QS $n$ examples stratified compared to the lowest QS $n$ examples stratified. Stratification was again applied to ensure the proportion of referable GSR was identical to that of the entire set. At low sample sizes, the delta is quite large, demonstrating the vast difference in quality and the ultimate impact these labels have on the model. As larger sample sizes are taken (including to the point of overlap), the delta closes, particularly as better examples are introduced for the low QS subset and worse examples are included for the high QS subset. 

\begin{table}[htb]
  \caption{AUC comparison of high QS and low QS for $n$ examples. SNCV was applied to determine these subsections of data and models were trained. The difference in AUC for top and bottom $n$ is shown below.}
  \label{tab:auc_highqs_stratified} 
  \begin{tabular}{ccll}
    \toprule
    Number of examples & High-quality & Low-quality & Delta\\
    \midrule
    1k & 0.848 & 0.301 & 0.547\\
    10k & 0.917 & 0.188 & 0.729\\
    25k & 0.940 & 0.212 & 0.728\\
    30k & 0.950 & 0.369 & 0.581\\
    35k & 0.951 & 0.600 & 0.351\\
    40k & 0.951 & 0.876 & 0.075\\
    45k & 0.949 & 0.927 & 0.023\\
    50k & 0.953 & 0.944 & 0.009\\
    55k & 0.953 & 0.952 & 0.001\\
    60k & 0.953 & 0.957 & -0.004\\
    65k & 0.948 & 0.955 & -0.006\\
    70k (All) & 0.954 & 0.954 & 0\\
  \bottomrule
\end{tabular}
\end{table}

From the highest quality 30k stratified train data, an AUC of 0.950 is obtained, showing little improvement in the additional 40k train data needed to reach the full train set performance at AUC of 0.954. While no performance improvements from sample selection were observed, these results confirmed that selecting examples with high-quality labels keeps enough examples required for training, as well as identifies the negative impact of low-quality labels on model performance. 

\subsection{Can SNCV be used to reduce the labeling requirements for a model?}

Next, we explored whether SNCV can be used to reduce the overall labeling requirements for training a deep learning model. We hypothesized that the effect of sample selection is more apparent when the total amount of labeled data is limited. In general, training with fewer examples results in poorer-performing models; under these conditions, the impact of noisy, low-quality labels may also be stronger. We asked: If our effort had started with a smaller total pool of labels, could SNCV be used to achieve similar performance to our model trained with more data? 

To assess the potential of SNCV to reduce labelling burden, we applied SNCV to a randomly sampled subset (40k subset) of total our full training dataset (`70k all'). We trained two additional deep learning models: one that used the entire set of 40k examples in the subset (`40k baseline'), and one that only used examples selected with a threshold quality score, after estimating quality using SNCV on the 40k subset. This high-quality subset incorporated the 30,000 examples with the highest QS, out of the pool of 40,000 (`40k SNCV'). The number 30,000 was determined from the tune performance between selecting 25,000, 30,000 and 35,000 examples. Both of these models were trained without using any information that might have been available with a larger labeling pool, mirroring a more resource-constrained effort. We assessed differences in performances using the DeLong test for correlated ROC curves \cite{delong1988comparing}, using a non-inferiority margin of 2\%. (We selected this margin \textit{a priori}, based on a consideration of recommendations for evaluating radiological models for medical devices \cite{gallas2012evaluating}, and consideration of the size of the test sets; we felt this provided a more stringent test than the 10\% margin recommended in \cite{gallas2012evaluating}, while a 1\% would be overly conservative given the sample sizes).

\begin{figure}[htb]
  \includegraphics[width=\linewidth]{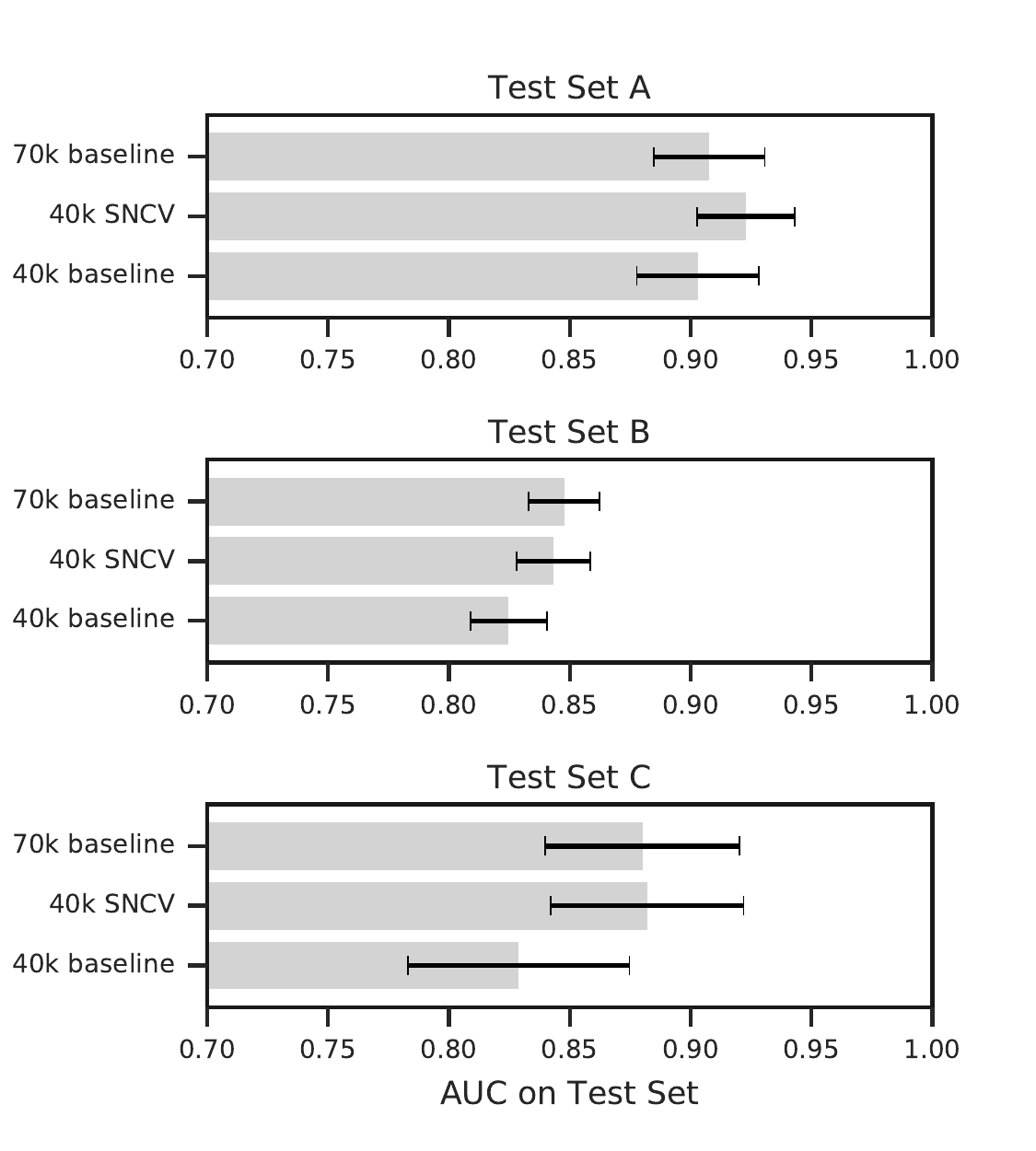}
  \caption{Summary of experiments using SNCV on a limited set of training data. Three separate models were
  trained on different sets of training data: ``70k baseline'' indicates the full set of 70,000 examples; ``40k SNCV'' indicates
  training with only selected high-quality-score examples from the 40k subset with stratification;
  ``40k baseline'' indicates a random subset of 40,000 examples, training with all the examples in the subset;  The subplots show performance of the three models
  on the three independent test sets used in \cite{phene2019deep}. Error bars denote confidence intervals determined by DeLong method.
  }
  \label{fig:labeling_strategies_summary}
\end{figure}

We evaluated performance of the three models on the three held-out test sets used in Phene et al \cite{phene2019deep}. Results are shown in Figure \ref{fig:labeling_strategies_summary}. As expected, when we train with fewer examples and make no other changes to the training set, the resulting model performs more poorly: AUC on the 40k all set is inferior to that on the 70k all set (compare first and third bars in each plot; two-sided DeLong test on nonequivalence, p = 0.008, p < 0.001, p < 0.001 for test sets A, B, and C). However, when we further restrict our 40k limited set to only include high-quality examples identified through SNCV, the resulting model performance increases, and approaches that of the model trained on the full 70k set (compare second bars in each subplot). Performance of the this model was non-inferior to the baseline model trained with the full dataset with margin of 2\% (Test set A: AUC 0.927 vs. 0.933, p = 0.003; Test set B: AUC 0.843 vs. 0.855, p = 0.003; Test set C: AUC 0.882 vs. 0.886, p = 0.007, DeLong test with a non-inferiority margin of 2\%). The AUC difference by applying SNCV to the 40k subset was statistically significant (Test set A: p = 0.029; Test set B: p < 0.001; Test set C: p < 0.001, Two-tailed DeLong test). These results show that SNCV can improve model performance when the amount of data is limited, lowering the labeling burden.  

\begin{table*}[htb]
  \caption{Delong test}
  \label{tab:delong}
  \begin{tabular}{p{3.2cm}p{4cm}p{4cm}p{4cm}}
    \toprule
    Null-hypotheses & 40k baseline inferior to 40k with SNCV by margin of 0.02 & 40k SNCV inferior to all 70k by margin of 0.02 & 40k baseline inferior to all 70k by margin of 0.02\\
    \midrule
    Test set A p-value & 0.496718 & 0.002697 (*) & 0.720615\\
    Test set B p-value & 0.370777 & 0.003483 (*) & 0.991617\\
    Test set C p-value & 0.990749 & 0.007238 (*) & 0.996128\\
    Results & - & Non-inferior & -\\
    \bottomrule
  \end{tabular}
\end{table*}

\subsection{Comparison between SNCV and NCV}
\begin{figure}[htb]
  \includegraphics[width=\linewidth]{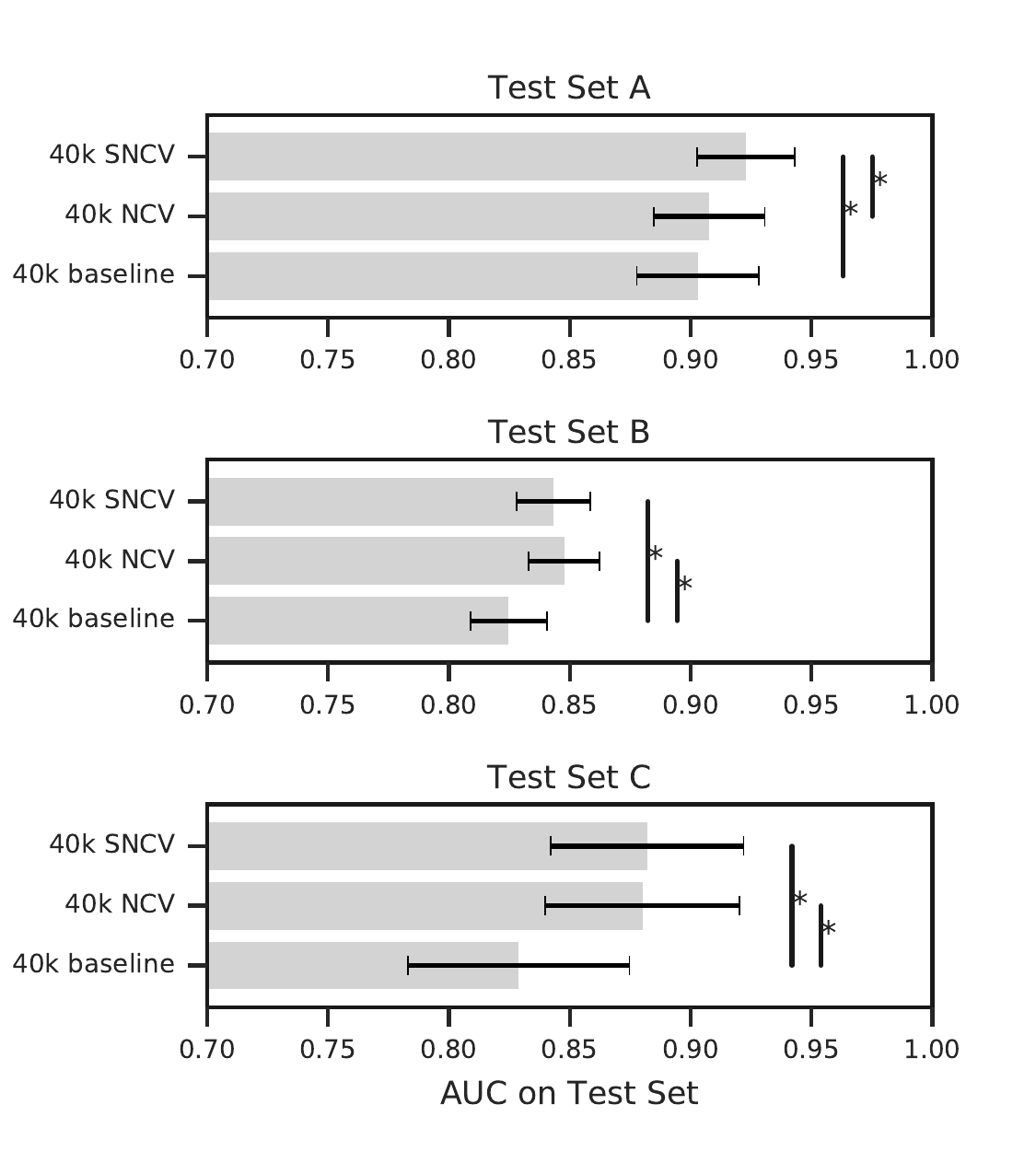}
  \caption{Summary of experiments comparing SNCV and NCV on a subset of 40k data. Three separate models were trained: all of the 40k, the 40k applying NCV and the 40k applying SNCV. The performance is plotted for three test sets. The vertical lines with * denote statistically significant differences (p<0.05, DeLong test).}
  \label{fig:sncv_ncv_comparison}
\end{figure}

Lastly, we directly compared the performance of the models trained with our SNCV method and the original NCV method \cite{chen2019understanding} (Figure \ref{fig:sncv_ncv_comparison}) applied to the same 40k subset used in the previous section. When the models were applied to Test Set A, SNCV showed statistically significant improvement over the baseline and NCV method. Wwhen the models were applied to Test Set B and C, both SNCV and NCV showed improvement over baseline, but there were not statistically significant differences between AUCs by the models trained with SNCV and NCV. The results suggest that sample selection based on cross validation can improve the model performance, and stratification based on labels with quality scores can make the method more effective in certain cases.

\section{Limitations and Future Work}

 Our study has several limitations. One limitation is that our current modeling approach is subject to variation due to randomness in model initialization and model training. Although we report confidence intervals through bootstrapping, this only captures variance in our test population, not the variance between models from repeated training runs. Future work could be to ensemble models these to mitigate the uncertainty during model training, or report the distribution across multiple training runs.
 
 Another limitation in our assessment of QS is using the argmax of the model rather than summing the softmax outputs for the overall referability. This has two consequences. First, the argmax of four tiers may not match the binary referral decision. For example, [0.05, 0.4, 0.3, 0.25] has an argmax of ``low risk'', which is no-refer, but summing the first two and last two classes would lead to a refer decision. Second, the QS is affected by how the probability is distributed across 4 classes. For example consider two images which have softmax probability output [0.5, 0.5, 0, 0] and [0.9, 0, 0.1, 0]. Our current method of selection would rank 0.9 higher than 0.5, but it does not capture the binary referral state which would be [1, 0] and [0.9, 0.1].

 One caveat of stratifying examples by labels given by graders is that the labels may have errors. Therefore the stratification is not based on the true classification, and in particular there is little justification to sample at the exact same proportion, because we do not know what the true ratio is between positive and negative cases. Instead of specifying the total number of samples to choose as a single number, we can specify how many positive examples and how many negative examples to choose separately, and examine what combination improves the model performance. Treating them as another set of hyper-parameters to tune is a potential extension of our method.
 
 We observed that the effectiveness of stratification depends on the test sets, and it is not clear when stratification is more effective. One hypothesis is that it is effective when the performance is validated by AUC with better ground-truth (e.g. by adjudication or by specialists). Hypothetically, if an image is initially graded by 10 times, and only flagged as positive in less than half, NCV without stratification may exclude all the positive labels. The two models can learn to predict that such images are more likely negative, and all the positive labels are considered incorrect and excluded. However, how likely an image is graded positive, even when below 50\%, can still be informative to predict the better ground-truth and improve AUC. Further analysis would help us understand how stratification helps, and will lead us to better applications of the method.

 One avenue of future work to consider is more experimentation with the methods of identifying correct examples. Chen et al \cite{chen2019understanding} used iterative NCV to select small number of examples and train a robust model on them before applying co-teaching. Using SNCV as an initial step may alleviate the issue of co-teaching that it can reinforce bias from initial training. Iterating more than once on SNCV or SNCV and co-teaching to include more examples is of future interest.

Using multifold SNCV or NCV could also be a future area of work to consider. The more data available for initial model training, the better the sample selection would be. Splitting the dataset in half has the limitation of only allowing each sub-model to learn from half of the total examples. Using larger fold (e.g. train a model using $90 \%$ of the data to select samples in the rest $10 \%$, and repeat this for 10 times) can potentially improve the performance of identifying actual errors and may improve the performance of the final model.

\section{Conclusions}
 We propose stratified noisy cross-validation, an extension of noisy cross validation\cite{chen2019understanding}, which allows us to measure confidence in model predictions and train models on selected examples with high quality labels. We also proposed a method for assigning quality scores to images and further verified that low quality score labels were likely due to grader errors. When a glaucoma specialist relabeled 1,277 of the lowest QS images, the doctor changed the GSR value for 85\% of them.
 
 In addition, when models were trained only with examples with low quality labels, the resultant model achieved less than 0.5 AUC on our tune set, indicating that the models were learning a negative correlation. Because our tune set was more robust (triple-graded by glaucoma specialists), it indicates that these low QS cases may in fact be  incorrect. In the medical domain, diagnoses can be ambiguous and label quality varies; our method helps identify high QS examples, in turn requiring fewer cases to be labeled to begin with.
 
 We also found ways to reduce labeling requirements for modeling efforts. Applying SNCV to roughly half our data, we were able to achieve non-inferior results to a full dataset. We found necessity in stratifying the examples to ensure inclusion of enough positive examples, because for some tasks (including diagnosing glaucoma), more challenging cases tend to be positives. Using only the model's softmax output as a quality score would indicate that uncertain cases would end up being lower quality, so we must enforce inclusion of enough positive cases for the model to learn from. Stratification helps handle class imbalance, which occurs commonly in medical imaging challenges.

 Our results show that this method can detect label errors in datasets and reduce labeling burden for model training on difficult medical imaging classifications. This could be especially helpful for work constrained in resources for data collection, because applying SNCV on a small amount of data can lead to equivalent performance as using a significantly larger set. Our SNCV method with quality score assignment can scale medical machine learning by making deep learning research more accessible with less labels needed for ambiguous medical tasks.


\bibliographystyle{ACM-Reference-Format}
\bibliography{main}

\appendix

\section{Supplementary Materials}

\subsection{Test sets}
Test set ``B'' comprised macula-centered color fundus images from the Atlanta Veterans Affairs (VA) Eye Clinic diabetic teleretinal screening program, which composed of 9,642 macula-centered color fundus images from the Atlanta VA Eye Clinic diabetic teleretinal screening program. Glaucoma-related International Classification of Diseases (ICD) codes on a per-patient basis and ONH referral codes recorded in the patient's medical history at any point in time before the capture of the image or up to 1 year after the image was taken were used as a reference standard for this dataset. Presence of such a code was considered ``referable'' because these are patients that were either referred for an in-person exam by the diabetic teleretinal screening program or were given such a code during an in-person exam. 

Test set ``C'' comprised macula-centered and disc-centered color fundus images from the glaucoma clinic at Dr. Shroff's Charity Eye Hospital, New Delhi, India. This dataset comprises 346 macula-centered and disc-centered fundus images from the glaucoma clinic at Dr. Shroff's Charity Eye Hospital. The reference standard for this dataset was an eye-level diagnosis of glaucoma as determined by a glaucoma specialist based on a full glaucoma workup that included history, clinical exam, visual field, and optical coherence tomography (OCT) assessment.

\subsection{Training hyperparameters}

We used the following hyperparameters for the Inception-v3 model:

\begin{verbatim}
Learning rate: 0.002
Batch size: 32
Data augmentation:
- Random horizontal reflections
- Random vertical reflections
- Random brightness changes 
  (with a max delta of 0.1147528)
  [see tf.image.random_brightness]
- Random saturation changes 
  between 0.5597273 and 1.27488455
  [see tf.image.random_saturation]
- Random hue changes 
  (with a max delta of 0.0251488) 
  [see tf.image.random_hue]
- Random contrast changes 
  between 0.9996807 and 1.7704824
  [see tf.image.random_constrast]
\end{verbatim}

\end{document}